%% file: main.tex
\pdfoutput=1
\documentclass[10pt,twocolumn,letterpaper]{article}
\usepackage{iccv}

\usepackage[dvipsnames, table]{xcolor}
\usepackage{times}
\usepackage{epsfig}
\usepackage{graphicx}
\usepackage{amsmath}
\usepackage{amssymb}
\usepackage{cite}
\usepackage{rotating}
\usepackage[ruled,linesnumbered]{algorithm2e}
\usepackage{bm}
\usepackage{algpseudocode}
\usepackage{booktabs}
\usepackage{lipsum}
\usepackage{enumitem}
\usepackage{algpseudocode} 
\usepackage{mathrsfs}
\usepackage{multirow}
\usepackage[most]{tcolorbox}
\usepackage{mathtools}
\usepackage{float}
\usepackage{soul}
\usepackage{caption}
\usepackage{subcaption} 
\usepackage[pagebackref=true,breaklinks=true,letterpaper=true,colorlinks,bookmarks=false]{hyperref}
\usepackage[capitalize]{cleveref}
\crefname{section}{Sec.}{Secs.}
\Crefname{section}{Section}{Sections}
\Crefname{table}{Table}{Tables}
\crefname{table}{Tab.}{Tabs.}

\input{notation}

\iccvfinalcopy %
\def\iccvPaperID{12654\vspace{-2.5mm}} %

\ificcvfinal\fi

\begin{document}

\title{SeeABLE: Soft Discrepancies and Bounded Contrastive Learning \\for Exposing Deepfakes\vspace{-4mm}}
\author{Nicolas Larue$^{1,2}$, Ngoc-Son Vu$^{1}$, Vitomir Struc$^{2}$, Peter Peer$^{2}$, Vassilis Christophides$^{1}$\\
  $^{1}$ETIS - CY Cergy Paris University, ENSEA, CNRS, France\\
  $^{2}$University of Ljubljana, Slovenia\vspace{-3mm}\\%
}

\maketitle
\ificcvfinal\thispagestyle{empty}\fi

\begin{abstract}\vspace{-3.5mm}
  Modern deepfake detectors have achieved encouraging results, when training and test images are drawn from the same data collection. However, when  these detectors are applied to images produced with unknown deepfake-generation techniques, considerable performance degradations are commonly observed. In this paper, we propose a novel deepfake detector, called SeeABLE, that formalizes the detection problem as a (one-class) out-of-distribution detection task and generalizes better to unseen deepfakes. Specifically, SeeABLE first generates local image perturbations (referred to as soft-discrepancies) and then pushes the perturbed faces towards predefined prototypes using a novel regression-based bounded contrastive loss. To strengthen the generalization performance of SeeABLE to unknown deepfake types, we generate a rich set of soft discrepancies and train the detector: (i) to localize, which part of the face was modified, and (ii) to identify the alteration type. To demonstrate the capabilities of SeeABLE, we perform rigorous experiments on several widely-used deepfake datasets and show that our model convincingly outperforms competing state-of-the-art detectors, while exhibiting highly encouraging generalization capabilities. The source code for SeeABLE is available from: {\small \url{https://github.com/anonymous-author-sub/seeable}}.
\end{abstract}
\vspace{-0.4cm}
\section{Introduction}
\label{sec:introduction}

Recent advances in (deep) generative models, such as generative adversarial networks (GAN) \cite{gan}, diffusion models \cite{denoisingdiffusionprobabilistic} and generative normalizing flows \cite{DensityestimationusingRealNVP}, have made it possible to generate fake images and videos with unprecedented levels of realism. Human faces have been a particularly popular target for such models, enabling the creation of so-called \textit{deepfakes} \cite{deepfakes, faceswap, face2face, neuraltextures}, i.e., manipulated facial images commonly used for malicious purposes. These deepfakes have been shown to constitute a serious psychological and financial treat to individuals, but also society as a whole \cite{harmpolitician, harmlooming}.
As a result, the deep learning community is actively working on countermeasures and detection techniques that can help to mitigate this threat.
\begin{figure}[!t]
  \centering
  \includegraphics[width=\linewidth]{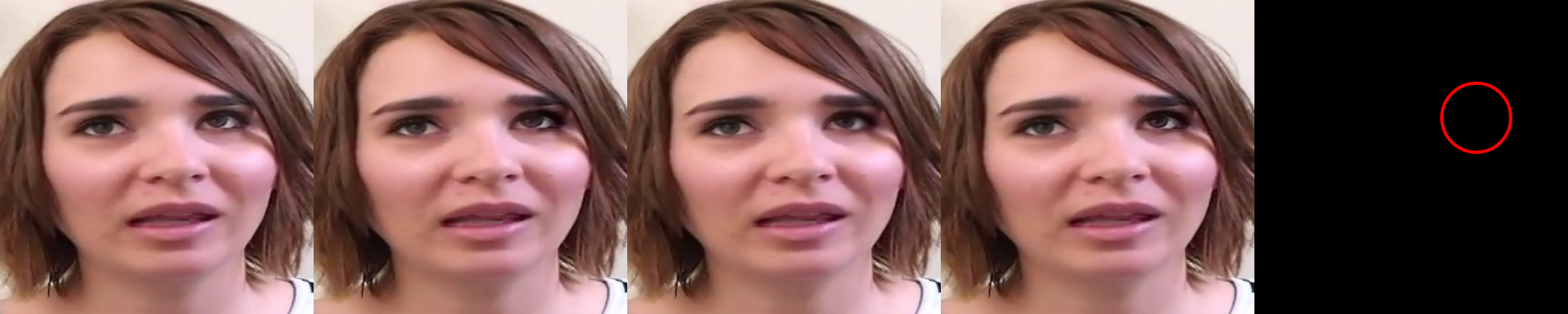}\vspace{-1mm}
  \caption{
    \textbf{Examples of faces with soft-discrepancies.}
    \textit{Can you identify the discrepancy in each image? SeeABLE can}. \textit{SeeABLE's answer: the perturbed area of the four facial images is within the circle shown on the right.} Note that a different soft discrepancy is used in each image.\vspace{-2mm}
  }
  \label{fig:discrepancies}
\end{figure}

By having access to datasets with both, real and forged (manipulated) faces \cite{celebDFv2, deeperforensics1, DeepfakeTIMIT, dfdcpreview, dfdc, FaceForensics++, UADFV}, existing deepfake detectors \cite{Mesonet, RecastingResidual, twobranch, xception, OntheDetection, FaceForensics++, dfpairwise} essentially learn a binary decision boundary that leads to reasonable detection performance with deepfake-generation techniques seen during training. However, as recent empirical studies \cite{LAE, FRETAL, DeepFakesandBeyond} report, the  performance of such (discriminatively-trained) detectors  degrades significantly when used with unseen face-manipulation methods, which severely limits their use in real-life deployment scenarios \cite{WildDeepfake}.
\begin{table*}[t!]
  \centering
  \resizebox{0.85\textwidth}{!}{%
    \begin{tabular}{l|cccc|cc|cc}
      \toprule
      \multicolumn{1}{l|}{\multirow{3}{*}{Deefake detection model}} & \multicolumn{4}{c|}{Augmentation (blended, adversarial)} & \multicolumn{2}{c|}{Alteration artifact} & \multicolumn{2}{c}{Problem formulation}                                                                                                                    \\
      \cmidrule{2-9}
      \multicolumn{1}{c|}{}                                         & \multicolumn{2}{c|}{global}                              & \multicolumn{2}{c|}{local}               & \multirow{2}{*}{global}                 & \multirow{2}{*}{local} & \multirow{2}{*}{classification} & \multirow{2}{*}{regression}                           \\
      \cmidrule{2-5}                                                & multiple                                                 & single                                   & multiple                                & single                 &                                 &                                                       \\
      \midrule
      Face Xray \cite{facexray}, PCL \cite{PCL}, OST \cite{ost}     & \checkmark                                               &                                          &                                         &                        & \checkmark                      &                             & \checkmark &            \\
      SLADD \cite{sladd}                                            &                                                          &                                          & \checkmark                              &                        & \checkmark                      &                             & \checkmark &            \\
      SBI \cite{sbi}                                                &                                                          & \checkmark                               &                                         &                        & \checkmark                      &                             & \checkmark &            \\ \midrule
      \textbf{SeeABLE}  (proposed)                                  &                                                          & \checkmark                               &                                         & \checkmark             &                                 & \checkmark                  &            & \checkmark \\ \bottomrule
    \end{tabular}}
  \caption{\textbf{Comparison of SeeABLE and SoTA detectors that use pseudo-deepfake synthesis during model learning.} The existing techniques differ in terms of augmentation techniques used, the level at which alterations are applied (local vs. global), and the problem formulation. As can be seen, SeeABLE differs significantly from existing techniques. \vspace{-1mm}%
  }
  \label{tab:differences}
\end{table*}

A powerful solution to improve the generalization capabilities of deepfake detectors is to use synthetic data (i.e., \textit{pseudo deepfakes}) during training and encourage the models to learn generalizable decision boundaries. Such strategies are at the core of many of the state-of-the-art (SoTA) detection models \cite{PCL,sbi,sladd,facexray,ost}
that either enrich the diversity of available deepfakes by synthesizing novel fake images for training, or rely completely on synthetic deepfakes when learning the detection models. These methods differ in the type of augmentations considered: blending- or adversarial-based techniques, adoption of global or local transformations, and use of single or multiple source/target images, as summarized in Table \ref{tab:differences}. Once the pseudo-fakes are generated, a classifier is learned to distinguish between real and fake faces. %
While these methods were observed to lead to highly competitive detection performance, especially in cross-dataset settings, they are still limited by the discriminative nature of the training procedure that tries to differentiate between real faces and the specific artifacts induced by the pseudo-deepfake generation procedure. %

In this paper, we propose a novel deepfake detector, called \textbf{SeeABLE} (\textbf{S}oft discr\textbf{e}panci\textbf{e}s \textbf{a}nd \textbf{b}ounded contrastive \textbf{l}arning for \textbf{e}xposing deepfakes), that formulates the detection problem as a (one-class) out-of-distribution detection task and generalizes better to unseen deepfakes than discriminatively-learned models. SeeABLE is trained  with images of real faces only and differs significantly from existing (pseudo-fake based) detectors, %
as seen in Table~\ref{tab:differences}. Specifically, the model first generates (subtle) local image perturbations, referred to as \textbf{{soft discrepancies}}, using a rich set of image transformations, as illustrated in Figure~\ref{fig:discrepancies}.  %
Next, the generated soft discrepancies are pushed towards a set of target representations (i.e., \textit{hard prototypes}) using a single \textbf{{multi-task regressor}} learned with a novel \textbf{bounded contrastive regression} loss. Here, the objective of the regressor is two-fold: (1) to map the different soft discrepancies into well-separated (and tightly clustered) prototypes that facilitate efficient similarity scoring (akin to prototype matching), and (2) to localize the spatial area of the local image perturbation and, thus, to exploit an auxiliary source of information for the regression task. The subtle image changes introduced by the soft discrepancies force SeeABLE to learn to detect minute image inconsistencies (and in turn a highly robust detector), whereas the local nature of the perturbation allows the model to exploit an additional localization (pretext) task that infuses complementary cues into the learning procedure. Unlike competing one-class detectors that typically rely on (low-level) per-pixel reconstructions to identify deepfakes, e.g. \cite{OCFakeDect}, SeeABLE, learns \textbf{rich} and \textbf{semantically meaningful features} for the detection process that, as we show in the experimental section, lead to highly competitive detection results. %

To demonstrate the capabilities of SeeABLE, we evaluate the model in comprehensive cross-dataset and cross-manipulation experiments on multiple datasets, i.e., FF++\cite{FaceForensics++}, CDF-v2 \cite{celebDFv2}, DFDC-p \cite{dfdcpreview} and DFDC \cite{dfdc}, and in comparison to twelve SoTa competitors. The results of the experiments show that the proposed model achieves highly competitive results on all considered datasets, while exhibiting encouraging generalization capabilities.

In summary, %
the main contributions of this paper are:
\begin{itemize}[noitemsep,leftmargin=*]\vspace{-1mm}
  \item We propose SeeABLE, a new state-of-the-art deepfake detector trained in one-class self-supervised anomaly detection setting that captures high-level semantic information for the detection task by localizing artificially-generated (spatial and frequency-domain) image perturbations. Unlike (most) competing solutions, SeeABLE learns to provide an anomaly score  that allows it to efficiently discriminate between real and fake imagery. %
  \item We introduce a novel Bounded Contrastive Regression (BCR) loss that enables SeeABLE to efficiently push/map the local soft discrepancies to a predefined set of (evenly-distributed) prototypes, and, in turn, to facilitate distance-based prototype matching for deepfake detection. %
  \item Through rigorous (cross-dataset and cross-manipulation) experiments on multiple dataset, we demonstrate the superior generalization capabilities of SeeABLE compared to existing (SoTA) deepfake detectors.
\end{itemize}

\section{Related work}
\label{sec:relatedwork}

In this section, we review (closely) related prior work needed to provide context for SeeABLE. %
For a more comprehensive coverage of the relevant topics, the reader is referred to some of the excellent surveys available \cite{mirsky2021creation,perera2021one}. %

\seg{Deepfake detection}
A considerable amount of deepfake detectors has been introduced in the literature over the years \cite{DeepFakesandBeyond, generalizedzeroandfew, beyondthespectrum, DetectingandRecoveringSequential,HierarchicalContrastiveInconsistency,ExplainingDeepfakeDetection}.
Early detectors, e.g., \cite{deepfakeopticalflowbased, DSPFWA, Mesonet, TowardsSolving, eyestellall}, relied mostly on the known deficiencies of deepfake-generation techniques and focused on the detection of the corresponding visual artifacts. %
A notable cross-section of these techniques \cite{FaceForensics++, SPSL, ThinkingInFrequency, secondorderlocalanomaly} use frequency-domain representations to discriminate between real and fake images.
Liu \etal \cite{SPSL}, for example, leveraged the phase spectrum to capture the up-sampling artifacts of face manipulation techniques, while
Qian \etal \cite{ThinkingInFrequency}, on the other hand, used a DCT-based model $F^3\text{net}$ \cite{F3Net} to extract frequency-domain cues and compute statistical features for forgery detection.
Fei \etal \cite{secondorderlocalanomaly} proposed a weakly supervised second order local anomaly learning module that decomposes local features by different directions and distances to calculate first and second order anomaly maps.
Another category of models \cite{delvingpatch, videoforensicsad, exploringtemporal, uiavit}  utilize temporal features for deepfake detection. The local and temporal-aware transformer-based deepfake detector (LTTD) \cite{delvingpatch}, for instance, utilized a local sequence transformer to model temporal consistency on restricted spatial regions to identify deepfakes. In \cite{videoforensicsad}, manipulated videos were detected based on the subtle inconsistencies between visual and audio signals by training an autoregressive model that captured the temporal synchronization between video frames and sound. Other recent works \cite{FRETAL, CDDB} also attempted to detect deepfakes in a continual learning setting, with the main goal of avoiding the catastrophic forgetting \cite{dsdm} across different tasks.

\seg{Detection through pseudo-deepfake generation}
One of the most effective strategies for learning deepfake
detectors that generalize well across deepfake generation techniques \cite{facexray, sbi, sladd, PCL, spatiotemporalregularity} is to use dedicated augmentation techniques to first synthesize forged images (i.e., pseudo deepfakes) and then train a binary classification model for the detection task.
The idea behind Face-Xray \cite{facexray}, for example, is to generate blended images (BI) of two different faces using a global transformation and then learn to discriminate between the real and blended faces. Shiohara \etal \cite{sbi} extended this idea and proposed a synthetic training dataset with self-blended images (SBI) that are generated from a single pristine/real face. SBI has been shown to generalize even better than Face-Xray to unseen deepfakes. SLADD \cite{sladd} used an adversarial training strategy to find the most difficult BI configuration and trained a classifier to predict the forgeries. In PCL \cite{PCL}, an inconsistency generator was used to synthesize forged data and patch-wise consistencies were later exploited to classify an image as either real or fake. %
Finally, OST \cite{ost} proposed a test-sample-specific auxiliary task, pseudo-training samples, and meta-learning to improve performance on identifying forgeries created by unseen methods.

\seg{One-class self-supervised anomaly detection (AD)}
AD, also referred to as out-of-distribution (OOD) detection, is an established research area, for which many techniques have been presented in the literature, including the one-class support vector machine (OC-SVM) \cite{OCSVM}, support vector data description (SVDD) \cite{SVDD}, deep OC-NN \cite{DeepOneClassNeuralNetwork} and multiple GAN-based methods \cite{YGan, GANomaly}. Very recently, self-supervised learning (SSL) has been successfully adopted for one-class AD \cite{MHROT, JezequelEfficient, csi}, with highly competitive results \cite{DeepOneClassNeuralNetwork,YGan,GANomaly}. In this setup, a (deep) model is commonly learned to solve an auxiliary (pretext) task in an SSL fashion. Then, to \textit{classify} a given input sample as either anomalous or normal, the result of the auxiliary task is evaluated with the provided test sample. Because the model is trained with normal data only, the assumption is that the model will perform well on normal data but fare poorly on anomalies. While different tasks were considered in the literature as pretexts \cite{MHROT,csi,cutpaste}, existing one-class AD models for deepfake detection, such as OC-FakeDect \cite{OCFakeDect}, relied exclusively on data reconstruction to facilitate the detection process.    %

\seg{Evenly-distributed prototypes}
\label{seg:pedcc}
Evenly-distributed points on a hypersphere maximize the average inter-class distance when being used as class centroids.
Formally, in a Euclidean space $\mathbb{R}^n$, a set of $K \in [2, \ldots, n+1]$ vectors $\{\boldsymbol{p}_1 \ldots \boldsymbol{p}_K\}$ vectors are evenly-distributed on a unit hypersphere such that,
$\forall\, i \neq j, \;  \boldsymbol{p}_{i} \cdot \boldsymbol{ p}_{j} = -1/(K-1)$.
When $K=n+1$, Lange and Wu~\cite{MMAlgo} proposed an analytical expression over the vertices of a regular simplex to generate $n$ points in $\mathbb{R}^{n}$.
Recently, \cite{dissectingsupcon} showed that contrastive learning objectives reach their minimum once the representations of each class collapse to the vertices of a regular simplex. %

\section{Methodology}
\label{sec:methodology}

\begin{figure*}
  \centering
  \begin{minipage}{.62\linewidth}
    \centering
    \includegraphics[height=5cm]{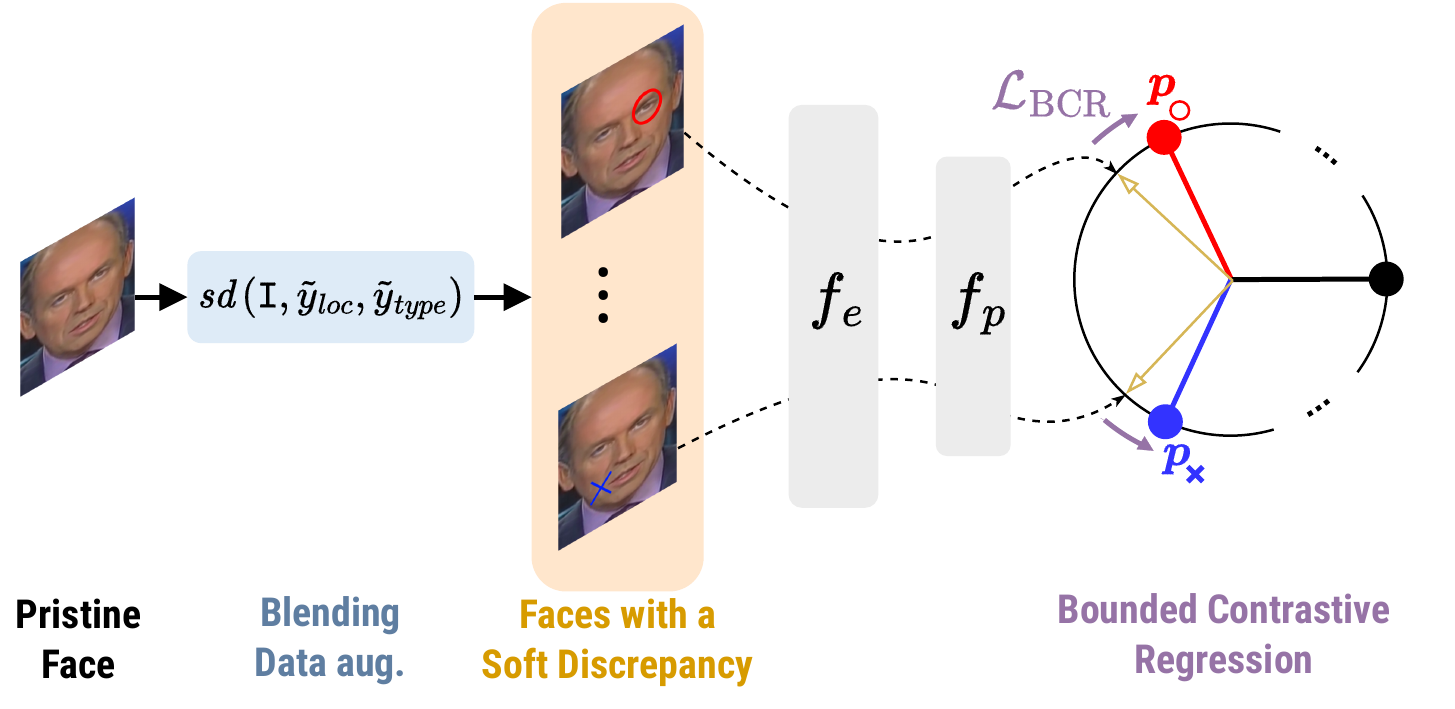}
    \caption*{(a) Training ($\notvector{p}_i$: predefined prototypes)}
    \label{fig:overview_a}
  \end{minipage}
  \hfill\vline\hfill
  \begin{minipage}{.37\linewidth} %
    \centering
    \includegraphics[height=5cm]{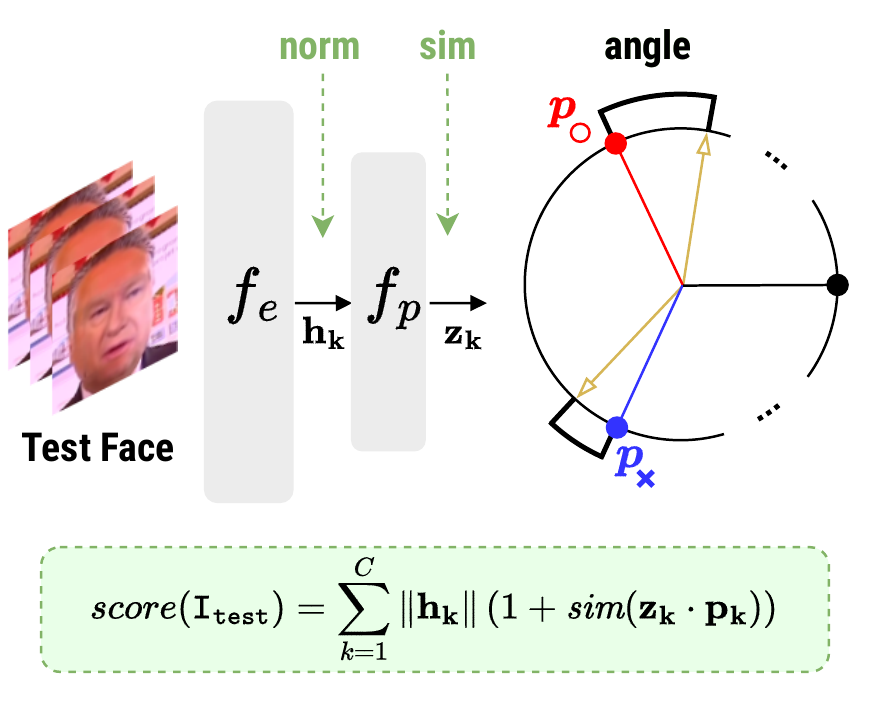}
    \caption*{(b) Anomaly detection}
    \label{fig:overview_b}
  \end{minipage}
  \caption{\textbf{High-level overview of SeeABLE.} The proposed deepfake detector is trained in a one-class self-supervised learning setting using real face images only. Once trained, SeeABLE is able to provide an  anomaly score that can be used for deepfake detection. $p_{\circ}$ and $p_{\times}$ are two different prototypes, the color encodes the position and the subscripts the discrepancy type.}\vspace{-2mm}
  \label{fig:overview}
\end{figure*}

Unlike competing methods that typically try to discriminate between real faces and (synthetically-generated) fakes, SeeABLE learns low-dimensional representations (hard prototypes) of synthetically-generated local image perturbations (soft discrepancies) and utilizes a prototype-matching procedure during inference to derive a (anomaly) score for the detection task. From a high-level  perspective, SeeABLE first uses a diverse set of augmentation techniques to create the soft discrepancies (\secref{sec:da}) and then trains a deep network to map them to a set of prototypes distributed evenly on a hypersphere, as also illustrated in Figure~\ref{fig:overview}. In the following sections, we detail both, %
the training (\secref{sec:learn}) and inference stages (\secref{methodo:inference}) of SeeABLE.

\subsection{Generating soft discrepancies (SD)}
\label{sec:da}
We use a parameterized data augmentation technique, denoted as $\notfunction{sd}$, to  create synthetic face images with subtle soft discrepancies (SDs).
Formally, given a dataset of real face images $\mathcal{D}_{real} = \ensemble{ {\notvector{I}_i} }_{i=1}^N$,
we generate a set of faces with synthesized soft discrepancies $\mathcal{D}_{sd}$ as follows:\vspace{-1mm}
\begin{equation}
  \mathcal{D}_{sd} = \ensemble{ ( \underbrace{\notfunction{sd}({\notvector{I}_i}, \notgt{{y_i}}_{loc}, \notgt{{y_i}}_{type})}_\text{Face with a soft discrepancy}, \underbrace{\notfunction{lbl}(\notgt{{y_i}}_{loc}, \notgt{{y_i}}_{type}}_\text{self-supervised signal} ) },   \vspace{-1mm}
\end{equation}
where ${\notvector{I}_i}$ is the $i$-th real image, $\notgt{{y_i}}_{type} \in \notintinter{N_{type}}$ denotes the type of soft discrepancy to be used for perturbing ${\notvector{I}_i}$, and $\notgt{{y_i}}_{loc} \in \notintinter{N_{loc}}$ stands for a label encoding the (discrete) location of the soft discrepancy.
Additionally, we use the generated ground truth, $\notgt{{y_i}}_{loc}$ and $\notgt{{y_i}}_{type}$, to construct a single label $\notgt{y_i} \in \notintinter{N_{loc} \times N_{type}}$ for each image in $\mathcal{D}_{sd}$ in a self-supervised fashion, as follows: \vspace{-1mm}
\begin{equation}
  \notgt{y_i}= \notfunction{lbl}(\notgt{{y_i}}_{loc}, \notgt{{y_i}}_{type})=\notgt{{y_i}}_{loc} \times N_{type} + \notgt{{y_i}}_{type}
  \label{eq:label}\vspace{-1mm}
\end{equation}
and $\notgt{{y_i}}_{loc}=\notfunction{pos}(\notgt{y_i})$ and $\notgt{{y_i}}_{type}=\notfunction{type}(\notgt{y_i})$ to invert the mapping. The presented process serves as a data factory and allows us to generate massive amounts of perturbed faces with corresponding labels for training the proposed model. %

\seg{a) Soft-discrepancy generation} SeeABLE uses a \textbf{blending} operation to generate locally perturbed facial images for training. %
Given a source and target face image ${\notvector{I}^s},{\notvector{I}^t} \in [0, 255]^{W\times H\times3}$, and a blending mask $\notvector{M} \in [0, 1]^{W \times H}$, the blending operation is defined as follows:\vspace{-1mm}
\begin{equation}
  \notfunction{blend} \left( \notvector{M}, {\notvector{I}^s}, {\notvector{I}^t} \right) = \notvector{M}
  \odot
  {\notvector{I}^s} +
  \left( 1-\notvector{M} \right)
  \odot
  {\notvector{I}^t},
  \label{eq:blending}
\end{equation}
where $\odot$ is the element-wise Hadamard product. %
Prior work \cite{facexray, sladd, sbi, spatiotemporalregularity} has adopted the above blending procedure to produce \emph{globally perturbed faces} that were then used to either fully substitute or to complement the dataset's actual deepfakes. Thus, the blending mask $\notvector{M}$ was defined in a way that affected the appearance of the entire face \cite{sladd}.
SeeABLE differs from these existing methods in that it generates faces with (subtle) local perturbations that not only encourage the model to learn a robust detector, but also enable using an additional localization task for the training procedure. To generate the locally perturbed images, we utilize \cref{eq:blending} with $N_{loc}$ different masks $\notvector{M}$ and $N_{type}$ different augmentations, which results in the following augmented dataset for the training of SeeABLE for each input image $\notvector{I}$:
\begin{equation}
  \notfunction{sd}(\notvector{I}, \notgt{{y}}_{loc}, \notgt{{y}}_{type})
  =
  \notfunction{blend} \left( \notfunction{Loc}_{}(\notvector{I},\notgt{{y}}_{loc}), \notvector{I}, \notfunction{Type}_{}(\notvector{I},\notgt{{y}}_{type}) \right),
  \label{eq:sd}
\end{equation}
where $\notfunction{Loc}$ is a function that generates a mask at location $\notgt{{y}}_{loc}$ and $\notfunction{Type}$ is a function that modifies the image using a specific augmentation technique defined by $\notgt{{y}}_{type}$.

\noindent \textbf{b) Discrepancy location.} %
To generate the local image perturbations with \cref{eq:sd}, SeeABLE requires suitable blending masks that ensure that only a specific region of the face is altered at the time. To this end, we define a \emph{submask scheme}, that partitions the facial are into a number of subcomponents. We use a simple \emph{grid} strategy for SeeABLE, where the facial region, defined by a set of facial landmarks, is divided into a grid with $N_{loc} = N_{rows} \times N_{cols}$ patches, where $N_{rows}$ and $N_{cols}$ represent the number of rows and columns in the grid, respectively (see Table \ref{tab:ablation_submask_properties}(d) for an illustration). The function $\notfunction{Loc}$, thus, returns a mask that corresponds to one of the $N_{loc}$ patches, defined by $\notgt{{y}}_{loc}$. We note that this strategy was found to work better than more complex semantics-driven submask schemes, and was, therefore, also used in the design of SeeABLE.   %

\noindent \textbf{c) Discrepancy type.} %
To detect object-level anomalies or out-of-distribution samples, self-supervised AD methods, such as \cite{csi, learningandevaluating}, use strong augmentations like rotations, patch permutations or cutpaste. %
Because the artifacts introduced by deepfake-generation techniques are typically subtle, we avoid such strong augmentations in SeeABLE and utilize more suitable perturbation techniques that generate soft (``not too pronounced'') discrepancies that enable our model to learn rich and robust features for the detection task. Specifically, we consider augmentations that yield discontinuities in both, the spatial and frequency domain. Thus, the number of different discrepancy types returned by the $\notfunction{Type}$ function from \cref{eq:sd}  is $N_{type}=2$ and is determined by $\notgt{{y}}_{type}$. Details of the concrete transformations used for the implementation are presented in Section~\ref{seg:implementation}.

\seg{d) Discrepancy invariant global transformations}
In the literature on self-supervised representation learning, special care is typically taken to avoid trivial solutions, \eg, when a network is trained to solve a pretext task to recognize the permutation of image patches and learns only the edge discontinuities instead of useful features. In SeeABLE, we avoid such trivial solutions (and over-fitting) by applying global \textit{discrepancy-invariant} transformations $\mathcal{T}_{inv}$ to $\notvector{I}$ before generating the local image perturbations. Because a family of global image transformations is used that do not impact the soft discrepancies, the model is encouraged to learn more general features that better capture the characteristics of the synthesized discrepancies.

\subsection{Learning the proposed model}
\label{sec:learn}

SeeABLE is trained in a multi-task fashion by minimizing the following learning objective:\vspace{-1mm}
\begin{equation}
  \mathcal{L}_{\text{SeeABLE}} = \notloss{BCR} + \lambda \, \notloss{GUI},\vspace{-1mm}
\end{equation}
where $\notloss{BCR}$ is the novel \textit{Bounded Contrastive Regression} (BCR) loss that aims to push the soft-discrepancies to the predefined hard prototypes, $\notloss{GUI}$ is the localization-related guidance loss that offers additional cues for the training procedure and infuses geometric constraints into the model optimization process, and  $\lambda$ is a balancing weight. \vspace{-1mm} %

\subsubsection{Supervised contrastive learning}
\label{sec:SupCon}
Assume a classification task with $K$ classes and a set of $N$ training samples $\{\left(\notvector{x}_i,\notgt{y_i}\right)\}_{i=1}^N$, %
where a (deep) encoder $\notfunction{f_e}$ first outputs an intermediate representation $\notvector{h}_i = \notfunction{f_e}(\notvector{x}_i)$ and the projector function $\notfunction{f_p}$, usually an MLP, then computes the final low-dimensional embedding $\notvector{z}_i = \notfunction{f_p}(\notvector{h}_i) \in \mathbb{R}^{D}$. %
Such models are often learned using \textbf{contrastive losses} that rely on the normalized temperature-scaled cross entropy (NT-Xent) \cite{simclr} optimization objective, defined for a positive pair of training examples $(\notvector{x}_i,\notvector{x}_p)$ as:\vspace{-1mm}
\begin{equation}
  \notloss{NT-Xent}\left(\notvector{z}_i, \notvector{z}_p\right) = -
  \log
  \frac{
  e^{ \notfunction{sim}( \notvector{z}_i, \notvector{z}_p ) / \tau}
  }{
  \sum^N_{\substack{j=1;~j\neq i}}
  e^{ \notfunction{sim}( \notvector{z}_i, \notvector{z}_j ) / \tau}
  },
  \label{eq:ntxent}
\end{equation}
where $\notfunction{sim}(\notvector{a},\notvector{b})=\notvector{a}\cdot\notvector{b}/(\norm{\notvector{a}} \norm{\notvector{b}})$ is the cosine similarity and
$\tau$ is a temperature variable. Recently, a novel loss was proposed in \cite{supcon} for fully supervised contrastive (SupCon) learning, where %
the training consists of two stages, where in \textbf{Stage 1} the encoder %
$\notfunction{f_e}$ and projector $\notfunction{f_p}$ are trained jointly to minimize:\vspace{-2mm}
\begin{equation}
  \notloss{SupCon} = \sum_{i=1}^{N} \frac{1}{|{P(i)}|} \sum_{p \in {P(i)}}
  \notloss{NT-Xent}\left(\notvector{z}_i, \notvector{z}_p\right),
  \label{eq:supcon}
\end{equation}
where
${P(i)}=\ensemble{p\in\notintinter{N} \backslash \ensemble{i} \,\lvert\, \notgt{y_p}=\notgt{y_i}}$ is a set of indices of the same label as $\notgt{y_i}$, and $\lvert \cdot \lvert$ is the set's cardinality, and in
\textbf{Stage 2} %
the encoder $\notfunction{f_e}$ is frozen and the projector $\notfunction{f_p}$ is replaced by a linear layer $\notfunction{f_{lin}}$, which is trained to solve the classification task using the standard cross-entropy loss.\vspace{-1mm}

\subsubsection{Bounded contrastive regression}
\label{sec:BCR}

One of the main contributions of this work is a novel loss for bounded contrastive regression (BCR). %
To introduce BCR, we first define a set $\ensemble{\notvector{p}_1, \ldots, \notvector{p}_K}$ of $K$ class prototypes and fix them to be \emph{evenly distributed} on a hypersphere in accordance with the procedure from \cite{MMAlgo}. Using these predetermined \textit{hard prototypes} as targets for the regression task, we then define the novel BCR loss as follows:\vspace{-1mm}
\begin{equation}
  \notloss{BCR} =
  \notloss{SupCon} +
  \sum_{i=1}^{N}
  \frac{
    \notloss{NT-Xent} \left(
    \notvector{z}_i,
    \notvector{p}_{\tilde{y}_{i}}
    \right)
  }{
    \abs{{P(i)}}
  },
  \label{eq:bcr}\vspace{-1mm}
\end{equation}
where $\mathcal{L}_{\text{SupCon}}$ and $\notloss{NT-Xent}$ are defined in \cref{eq:supcon} and \cref{eq:ntxent}.
As can be seen, computing the BCR involves creating positive pairs that combine the embeddings $\notvector{z}_i$ with their corresponding class prototypes $\notvector{p}_{\tilde{y}_{i}}$. Such positive pairs encourage the supervised contrastive loss to generate clusters around the class prototypes.
Our intuition here is that this will enable the regression model to generate bounded hypherspherical cap (evenly distributed and equally sized) clusters for each class within the unit-hypersphere.

\seg{BCR properties}
Given a sample $\left(\notvector{x}_i, \notgt{y_i} \right)$ and its embedding $\notvector{z}_i=\notfunction{f_p}(\notfunction{f_e}(\notvector{x}_i))$, BCR has the following properties:
\begin{itemize}[noitemsep,leftmargin=*]
  \item \textbf{Optimal representation.}
        Since we choose the prototypes to be optimal (evenly distributed) for contrastive losses, \cref{eq:bcr} tends to its minimum when all embeddings collapse to their corresponding class prototypes:
        \begin{equation}
          \forall i \in \notintinter{N} ~,~\notvector{z}_i = \notvector{p}_{\notgt{y}_i}
        \end{equation}

  \item \textbf{Regression.} By introducing a scalar variable $r_i \in \mathbb{R}^{+}$,
        for the cosine similarity between $\notvector{z}_i$ and $\notvector{p}_{\notgt{y_i}}$:
        \begin{equation}
          r_i = \notfunction{d_{sim}}( \notvector{z}_i, \notvector{p}_{\notgt{y_i}} ) = 1 - \notfunction{sim}(\notvector{z}_i, \notvector{p}_{\notgt{y_i}})
          \label{eq:bcr_regression}
        \end{equation}
        we observe that \cref{eq:bcr} implicitly regress $r_i$ to zero ($r_i \rightarrow  0$), with the equality to zero ($r_i =  0)$ occurring with the optimal representation.

  \item \textbf{Efficient prediction.} Given an embedding $\notvector{z}_i$, we can obtain its prediction (i.e., location and discrepancy type) through an efficient prototype-matching procedure, i.e.:
        \begin{equation}
          y_i = \argmin_{k} \ensemble{
            \notfunction{sim}\left(
            \notvector{z}_i, \notvector{p}_k
            \right)
            \vert\
            k \in \notintinter{K}
          }
          \label{eq:prediction}
        \end{equation}\vspace{-3mm}
\end{itemize}

\subsubsection{Guidance loss}
\label{sec:GUI}

In order to further enhance the performance of SeeABLE, we introduce an additional guidance loss for the regression task that incorporates task-specific knowledge.
To illustrate the idea, let us assume that the model produces a completely erroneous prediction instead of the expected ground truth, i.e., $y \neq \notgt{y_i}$,
Since our the prototypes are distributed evenly, the distance to the incorrect class is always the same, regardless of error, i.e., $\notfunction{d_{sim}}(\notvector{z}_i, \notvector{p}_{\notgt{y_i}}) = K/(K-1)$, resulting in an \emph{equal treatment} of all prediction errors.
To address this issue, we propose a loss that explicitly weights the distances $r_i$ based on some prior knowledge encoded in $G$: %
\begin{equation}
  \notloss{GUI} = \sum_{i \in [1 \isep N]}{ \notfunction{G}\left(y_i, \notgt{y}_{i}\right) \times r_i }.
\end{equation}

\seg{Model guidance with geometric constraints}
In SeeABLE, we consider facial geometry and geometric constraints as sources of prior knowledge for $\notfunction{G}$ and incorporate these into the discrepancy localization objective of the proposed detector. The main idea here is to use lower penalties for errors that originate from facial symmetry (e.g., substitutions of the left for the right eye) than other types of localization errors. Similarly, mispredictions of the type of soft-discrepancy within a region should be penalized less than mispredictions of the discrepancy location. %
Formally, $\notfunction{G}$ is defined as follows:
\begin{equation}
  \notfunction{G}(y_i, \notgt{y}_{i}) =
  \begin{cases}
    2^{-2}  \qquad\qquad\qquad\quad\,\:\:\: \textcolor{gray}{\text{if } \notfunction{pos}(y_i) = \notfunction{pos}(\notgt{y}_{i})}         \\
    2^{-1}  \qquad\quad~\:  \textcolor{gray}{\text{else if } \notfunction{sym}(\notfunction{pos}(y_i)) = \notfunction{pos}(\notgt{y}_{i})} \\
    2^{-0} \times d_{graph}(\notfunction{pos}(y_i), \notfunction{pos}(\notgt{y}_{i})) \quad\textcolor{gray}{\text{otherwise}}
  \end{cases}
\end{equation}
where $\notfunction{sym}(\cdot)$ returns the location of symmetric patch w.r.t. the vertical center axis of the image and $d_{graph}$ is a graph-based distance between the location of two patches.

\subsection{Inference: anomaly score computation}
\label{methodo:inference}
SeeABLE utilizes a deepfake detection score that is derived from the cosine similarity (\cref{eq:bcr_regression}) with the trained prototypes, serving as a direct cue. Moreover, the norm is employed as an indirect cue of the model's confidence.%
We, therefore, define the anomaly score for a test image $\notvector{I}_{\text{test}}$ as:
\begin{equation}
  \notfunction{score}(\notvector{I}_{\text{test}})=
  \sum_{k =1}^K \,
  \underbrace{
    \norm{ \notvector{h}_k }
  }_{\text{indirect}}
  \times
  \underbrace{\left(
    1 + \notfunction{sim}
    \left(
    \notvector{z}_k,
    \notvector{p}_k
    \right)
    \right)}_{\text{direct}}
  \label{eq:score}
\end{equation}
where $\notvector{h}_k = f_{e}( \notfunction{sd}( \notvector{I}_{\text{test}}, \notfunction{pos}(k), \notfunction{type}(k) ) )$ and  $\notvector{z}_k = f_p(\notvector{h}_k) / \Vert{f_p(\notvector{h}_k)} \Vert$.
Note that $1+\notfunction{sim}(\notvector{z}_k, \notvector{p}_k) = 2 - d_{sim}( \notvector{z}_{k}, \notvector{p}_k) \ge 0$.

\section{Experiments}
\label{sec:experiments}

\subsection{Implementation details}\label{seg:implementation}

\noindent \textbf{Model implementation.}
As in \cite{sbi}, we use EfficientNet-b4 \cite{efficientnet} as the encoder $f_e$ of SeeABLE, and a one-layer MLP (with $D=128$ outputs) followed by a $\ell_2$ normalization for the projection layer $f_p$.
The number of vertices (prototypes) of the regular simplex in $\mathbb{R}^{K-1}$ is set to $K=33$. SeeABLE is trained for $200$ epochs with the SGD optimizer, a batch size of $6$ and a learning rate of $1e^{-3}$ that is decayed to $1e^{-5}$ with a cosine scheduler. During training, the balancing parameter $\lambda$ is first set to $0$ and then increased gradually to $0.1$ to strengthen the importance of the geometric constraint in later epochs. Once trained, SeeABLE requires around $15$ ms to process one frame on a PC with
an RTX $3060$. %

\seg{Considered transformations} For the global transformations $\mathcal{T}_{inv}$, we consider the following operations: (1) random translations of up to $3\%$ and $1.5\%$ of the image width and height, respectively, (2) random scaling (followed by center cropping) by up to $5\%$, and (3) random shifting of HSV channel values by up to $0.1$.
To generate the \textit{SDs in the spatial domain}, we use: (1) shifting of RGB channel values by up to $20$, (2) random shifting of HSV channel values by up to $0.3$, and (3) random scaling of the brightness and contrast by a factor of up to $0.1$. For the \textit{SDs in the frequency domain}, one of the following operators is used: (1) down-sampling by a factor of $2$ or $4$, (2) application of a sharpening filter and blending with the original with an $\alpha$ value in the range $[0.2, 0.5]$, and (3) JPEG compression with a quality factor between $30$ and $70$. These hyperparameter values were selected in a way that resulted in visually subtle soft discrepancies without major artifacts, similarly to~\cite{sbi}.

\seg{Data preprocessing}
We use a pretrained RetinaFace \cite{retinaface} model for face detection and dlib \cite{dlib} for locating $68$ facial landmarks. The detected face regions are resized to $256\times256$ pixels (using bilinear interpolation) prior to the experiments. No special effort is made to align the faces across frames and the landmarks are used only to define the region-of-interest and the grid for the guidance loss.

\seg{Deepfake detection} During the inference phase, we uniformly sample $30$ frames per test video. The anomaly score is first computed for each frame separately using \cref{eq:score}, and the anomaly score for the entire video is obtained by averaging the frame-level scores.

\subsection{Experimental setup}

\noindent \textbf{Experimental datasets.}
As in \cite{FaceForensics++, csi, sladd, PCL}, we use the \textit{FaceForensics++} (FF++) dataset to train our model. The FF++ dataset \cite{FaceForensics++} contains $1000$ videos that are split into three groups: $720$ videos for training, $140$ for validation and $140$ for testing. We utilize only the pristine training videos for learning SeeABLE and sample at most $1$ frame per video when constructing batches for the optimization procedure. To evaluate SeeABLE in cross-manipulation settings, we adopt the deepfake (test) part of FF++, where each video is generated using one of the four deepfake-generation techniques: Deepfakes (DF)\cite{deepfakes}, Face2Face (F2F) \cite{face2face}, NeuralTextures (NT) \cite{neuraltextures} and FaceSwap (FS) \cite{faceswap}.

To demonstrate the performance of SeeABLE in cross-dataset settings, three additional datasets are adopted, i.e., Celeb-DF-v2 (CDF-v2) \cite{celebDFv2}, DeepFake Detection Challenge preview (DFDC-p), and DeepFake Detection Challenge public (DFDC) \cite{dfdc}. The CDF-v2 dataset \cite{celebDFv2} consists of $590$ real and $5,639$ deepfake celebrity videos, generated using a sophisticated deepfake approach. The DFDCp dataset contains over $5000$ videos (original and fake) and features two deepfake generation methods, while the DFDC dataset \cite{dfdc} comprises over $128,000$ video sequences with more than $100,000$ deepfakes of different quality. %

\seg{Performance indicators} In line with standard eval\-uation methodology \cite{sbi,sladd, PCL}, we use the Area Under the Receiver Operating Characteristic Curve (AUC) to evaluate the performance of our detector in a threshold-free manner. %

\seg{SoTa baselines} We compare SeeABLE to multiple SoTA competitors: (1) \textit{pseudo-deepfake} based detection methods, i.e., DSP-FWA \cite{DSPFWA}, Face X-ray \cite{facexray}, SLADD \cite{sladd}, PCL \cite{PCL}, SBI \cite{sbi}, and OST \cite{ost} (2) \textit{video-based techniques}, i.e., Two-branch \cite{twobranch} and LipForensics \cite{LipForensics}, (3) \textit{transformer-based} methods, i.e., UIA-ViT \cite{uiavit}, FTCN-TT \cite{exploringtemporal}, and LTTD \cite{delvingpatch}, and (4) two versions of the \textit{one-class} OC-FakeDect model \cite{OCFakeDect}. For a fair comparison, results from the original papers are reported (where available).

\begin{table}[t!]
  \centering
  \resizebox{\columnwidth}{!}{%
    \begin{tabular}{lc|ccccc}
      \toprule
      \multirow{2}{*}{Method}          & \multirow{2}{*}{Pristine \newline} & \multicolumn{4}{c}{Test set - AUC (\%)}                                                 \\
      \cmidrule{3-6}
                                       & only                               & CDFv2                                   & DFDC          & DFDCp         & Avg.          \\
      \midrule

      DSP-FWA \cite{DSPFWA}            & \checkmark                         & 69.3                                    & -             & -             & 69.3          \\
      Two-branch \cite{twobranch}      &                                    & 76.6                                    & -             & -             & 76.6          \\
      LipForensics \cite{LipForensics} &                                    & 82.4                                    & 73.5          & -             & 77.9          \\
      Face X-ray \cite{facexray}       &                                    & 79.5                                    & 65.5          & -             & 72.5          \\
      SLADD \cite{sladd}               &                                    & 79.7                                    & -             & 76.0          & 77.8          \\
      PCL+I2G \cite{PCL}               & \checkmark                         & \textbf{90.0}                           & 67.5          & 74.4          & 77.3          \\
      SBI$^\dagger$ \cite{sbi} %
                                       & \checkmark                         & 85.9                                    & 69.8          & 74.9          & 76.9          \\
      OST \cite{ost}                   &                                    & 74.8                                    & -             & 83.3          & 79.1          \\
      UIA-ViT \cite{uiavit}            & \checkmark                         & 82.4                                    & -             & 75.8          & 79.1          \\
      FTCN-TT \cite{exploringtemporal} &                                    & 86.9                                    & 74.0          & -             & 80.4          \\
      LTTD \cite{delvingpatch}         & \checkmark                         & 89.3                                    & -             & 80.4          & -             \\

      \midrule
      \textbf{SeeABLE} (ours)          & \checkmark                         & 87.3                                    & \textbf{75.9} & \textbf{86.3} & \textbf{83.2} \\
      \bottomrule
      \multicolumn{6}{l}{$^\dagger$SBI was re-evaluated using the official code with $\textsc{M}_{\text{ConvexHull}}$.}                                               \\
    \end{tabular}}\vspace{-2mm}
  \caption{\textbf{Comparison of SeeABLE and SoTA methods in the cross-dataset scenario.} For a fair comparison, the reported results are  cited directly from the original papers. %
    \vspace{-1.5mm}}
  \label{tab:exp_extra}
\end{table}

\subsection{Performance evaluation}

\seg{Cross-dataset evaluation} Table \ref{tab:exp_extra} compares the performance of SeeABLE with SoTA detectors in a cross-dataset scenario. Here, all models are trained on FF++ and tested on datasets not seen during training. As can be observed, SeeABLE yields the best overall (average) performance, while being among the simplest of all considered detectors. Unlike other competitors, the proposed model does not rely on adversarial training schemes or availability of deepfake examples during training, but still leads to highly competitive results. SeeABLE convincingly outperforms all methods on the DFDC and DFDCp datasets, including all pseudo-deepfake based detectors, video-based models, one-shot techniques and transformer-based models, while being rivaled on CDFv2 only by the video transformer based detector LTTD \cite{delvingpatch} and the PCL-I2G technique \cite{PCL}.
For a fair interpretation of the results, it should be noted that SBI was evaluated using the official code associated with $\textsc{M}_{\text{ConvexHull}}$ (see Table~\ref{tab:ablation_submask_properties}(a)) since the results reported in the original paper were obtained with a more sophisticated masking scheme and a Sharpness Aware Minimization (SAM) \cite{sam} procedure, which commonly improves performance at the cost of doubling the training time. We also note that some of the competitors do not provide results for all test datasets, but SeeABLE is still the top performer  even if the average AUC scores are computed only across the available results. For instance, the average AUC of SeeABLE and LTTD on CDFv2 and DFDCp are 86.8\% and 84.8\%, respectively.

\begin{table}[t]
  \centering
  \resizebox{0.84\columnwidth}{!}{%
    \begin{tabular}{l|cccc|c}
      \toprule
      \multirow{2}{*}{Method}            & \multicolumn{5}{c}{Test set - AUC (\%)}                                      \\
      \cmidrule { 2 - 6 }                & DF                                      & F2F  & FS   & NT   & Avg.          \\
      \midrule
      OC-FD1$^\dagger$ \cite{OCFakeDect} & 86.2                                    & 70.7 & 84.8 & 95.3 & 84.2          \\
      OC-FD2$^\dagger$ \cite{OCFakeDect} & 88.4                                    & 71.2 & 86.1 & 97.5 & 85.8          \\
      Face X-ray \cite{facexray}         & -                                       & -    & -    & -    & 87.3          \\
      SBI \cite{sbi}                     & 97.5                                    & 89.0 & 96.4 & 82.8 & 91.4          \\
      OST \cite{ost}                     & -                                       & -    & -    & -    & 98.2          \\
      SLADD \cite{sladd}                 & -                                       & -    & -    & -    & 98.4          \\ \midrule

      \textbf{SeeABLE}  (ours)           & 99.2                                    & 98.8 & 99.1 & 96.9 & \textbf{98.5} \\
      \bottomrule
      \multicolumn{6}{l}{\footnotesize$^\dagger$OC-FD1 and OC-FD1 refers to two versions of OC-FakeDect}
    \end{tabular}}\vspace{-1.5mm}
  \caption{
    \textbf{Cross-manipulation evaluation on FF++ HQ.} SeeABLE achieves SoTA results on all subsets of FF++. SLADD, OST, Face Xray report only the average result.}\vspace{-2mm}
  \label{tab:exp_intra_c23}
\end{table}

\seg{Cross-manipulation evaluation} A key aspect of deepfake detectors is their generalization to different manipulation techniques. Following the evaluation protocol from \cite{PCL}, we evaluated SeeABLE on the four manipulation methods of FF++, i.e., DF, F2F, FS, and NT.
As in all experiments presented in this paper, the raw version of FF++ is used for training and the HQ version is considered for testing. As can be seen from Table \ref{tab:exp_intra_c23}, SeeABLE outperforms all competing detectors on all four manipulation types with an average AUC of 98.5\%. Especially interesting here is the comparison to SeeABLE's closest competitors, the one-class OC-FakeDect detectors, which the proposed model outperforms by a margin of more than $12\%$.

\subsection{Ablation study}

\seg{Backbone impact}
In Table \ref{tab:ablation_backbones}, we evaluate the effect of different (backbone) encoder architectures $f_e$ on the performance of SeeABLE, i.e., ResNet-50 \cite{resnet}, Xception \cite{xception} and EfficientNet-b4. We observe that the best overall performance is obtained with EfficientNet-b4, which outperforms the runner-up, Xception \cite{xception}, by 3.5\%, and the weakest model (ResNet-50) by 3.8\%. In general, larger and more powerful encoders lead to better generalization, but even with the weaker backbones, SeeABLE still outperforms many of the SoTA competitors from Table \ref{tab:exp_intra_c23}. These results suggest that SeeABLE is applicable to different backbone models and is expected to further benefit from future developments in model topologies.    %

\begin{table}[t]
  \centering
  \resizebox{0.93\columnwidth}{!}{%
    \begin{tabular}{l|cccc|c}
      \toprule
      \multirow{2}{*}{Encoder ($f_e$)}    & \multicolumn{5}{c}{Test set - AUC (\%)}                                                                 \\
      \cmidrule { 2 - 6 }                 & DF                                      & F2F           & FS            & NT            & Avg.          \\
      \midrule
      ResNet50 \cite{resnet}              & 96.2                                    & 94.7          & 95.2          & 92.6          & 94.7          \\
      Xception \cite{xception}            & 94.5                                    & 95.0          & 96.4          & 94.2          & 95.0          \\
      EfficientNet-b4 \cite{efficientnet} & \textbf{99.2}                           & \textbf{98.8} & \textbf{99.1} & \textbf{96.9} & \textbf{98.5} \\
      \bottomrule
    \end{tabular}}\vspace{-1mm}
  \caption{\textbf{Performance of SeeABLE with different backbones.} Results are shown in terms of AUC (in \%) on FF++.}\vspace{-2.5mm}
  \label{tab:ablation_backbones}
\end{table}

\seg{Regression vs. classification}
We explore in Table \ref{tab:ablation_loss} the impact of learning SeeABLE within the proposed regression task (using $\notloss{BCR}$), as opposed to classification tasks (learned with cross-entropy  $\notloss{CE}$ and supervised-contrastive $\notloss{SupCon}$ losses). For a fair comparison, we compare all strategies with and without the geometric constraint ($\notloss{GUI}$) and the best performing hyperparameters.

As can be seen, $\notloss{BCR}$ convincingly outperforms $\notloss{CE}$ and $\notloss{SupCon}$ with a performance difference of 7.3\% and 4.3\%.
When adding the geometric constraint $\notloss{GUI}$, an additional improvement of 2.2\% for $\notloss{BCR}$, 1.7\% for $\notloss{CE}$, and 1.9\% for $\notloss{SupCon}$ can be observed.
This illustrates the strength of the proposed regression-based learning objective.
We conjecture that the drop in detection performance when changing from the regression loss to the classification losses is due to the fact that: (1) the representation learned through $\notloss{CE}$ and $\notloss{SupCon}$ is not optimized for the cosine-similarity-based scoring; (2) some useful characteristics, such as the hardness-aware property \cite{UnderstandingContrastiveLoss} of the contrastive loss, were lost, ultimately leading to suboptimal detection results.

\begin{table}[t]
  \centering
  \resizebox{0.79\columnwidth}{!}{%
    \begin{tabular}{l|cccc|c}
      \toprule
      \multirow{2}{*}{Method}              & \multicolumn{5}{c}{Test set - AUC (\%)}                                                                 \\
      \cmidrule { 2 - 6 }                  & DF                                      & F2F           & FS            & NT            & Avg.          \\
      \midrule
      $\notloss{CE}$                       & 96.8                                    & 91.7          & 86.7          & 80.8          & 89.0          \\
      $\notloss{CE}$ + $\notloss{GUI}$     & 96.3                                    & 92.6          & 89.8          & 84.0          & 90.7          \\
      \midrule
      $\notloss{SupCon}$                   & 96.9                                    & 94.4          & 91.3          & 85.4          & 92.0          \\
      $\notloss{SupCon}$ + $\notloss{GUI}$ & 97.9                                    & 97.1          & 93.1          & 87.4          & 93.9          \\
      \midrule

      $\notloss{BCR}$                      & 97.4                                    & 96.1          & 96.4          & 95.5          & 96.3          \\
      $\notloss{BCR}$ +  $\notloss{GUI}$   & \textbf{99.2}                           & \textbf{98.8} & \textbf{99.1} & \textbf{96.9} & \textbf{98.5} \\
      \bottomrule
    \end{tabular}}\vspace{-1mm}
  \caption{\textbf{Learning in a regression vs. classification settings.} Shown are AUC scores (in \%) on FF++.\vspace{-1mm}}
  \label{tab:ablation_loss}
\end{table}
\begin{table}[t]
  \centering
  \resizebox{0.9\columnwidth}{!}{%
    \begin{tabular}{ccc|cccc|r}
      \toprule
      $\notloss{BCR}$ & $\notloss{GUI}$ & $\lambda$  & DF            & F2F           & FS            & NT            & Avg.          \\
      \midrule
      \checkmark      & -               & -          & 97.3          & 96.1          & 96.3          & 95.4          & 96.3          \\
      -               & \checkmark      & -          & 90.3          & 87.0          & 87.9          & 82.0          & 86.8          \\
      \checkmark      & \checkmark      & const.     & 97.8          & 96.0          & 97.1          & 94.4          & 96.4          \\
      \checkmark      & \checkmark      & $\searrow$ & 98.7          & 98.7          & 95.4          & 97.8          & 97.6          \\
      \checkmark      & \checkmark      & $\nearrow$ & \textbf{99.2} & \textbf{98.8} & \textbf{99.1} & \textbf{96.9} & \textbf{98.5} \\
      \bottomrule
    \end{tabular}}\vspace{-1mm}
  \caption{
    \textbf{Impact of loss terms ($\notloss{BCR}$, $\notloss{GUI}$) and balancing strategies ($\lambda$) on performance (AUC in \%) on FF++.}\vspace{-3mm}}
  \label{tab:exp_effectiveness}
\end{table}

\seg{Contribution of different losses}
\label{ablation:_multiobjective}
In Table \ref{tab:exp_effectiveness}, we investigate the impact of the two loss terms utilized to learn SeeABLE, i.e., $\notloss{BCR}$ and $\notloss{GUI}$, as well as the strategy used to balance the two during training. %
Several cases are considered: (1) a fixed trade-off with $\lambda=\text{const.} = 0.1$, (2) increasing $\lambda$ linearly from $0$ to $0.1$ during training - marked $\lambda=\nearrow$, (3) decreasing  $\lambda$ linearly from 0.1 to 0 during training - marked $\lambda=\searrow$. In general, we see that $\notloss{BCR}$ and $\notloss{GUI}$ complement each other and better results are obtained when both are considered jointly, as opposed to either one alone. Additionally, we observe that the best overall performance is achieved when the importance of the geometric constraint is gradually increased during training. %

\seg{Effect of submask generation strategies}
\label{ablation:face_partitioning}
In Table~\ref{tab:ablation_submask_properties},~we evaluate the impact of different submask-generation strategies on the performance of SeeABLE on the DFDC dataset. Specifically, we consider: (1) the single global mask strategy $\textsc{M}_{\text{ConvexHull}}$ used in \cite{facexray, sbi}, (2) the semantics-guided strategy $\textsc{SM}_{\text{SLADD}}$ used in SLADD \cite{sladd}, a baseline meshgrid strategy $\textsc{SM}_{\text{Meshgrid}}$ and the $4\times 4$ patch-based strategy $\textsc{SM}_{\text{Grid}}$ used with SeeABLE. The four strategies are illustrated in Table~\ref{tab:ablation_submask_properties}(a)-(d). As can be seen, the best performance is obtained with $\textsc{SM}_{\text{grid}}$ with 4 rows and 4 columns. In $\textsc{SM}_{\text{Grid 4x4}}$, $33$ hard prototypes are used ($C=1+2\times4\times4$). We note that the similar results are obtained with $\textsc{SM}_{\text{Grid 3x3}}$ and $\textsc{SM}_{\text{Grid 5x5}}$. In SeeABLE, a good submask-generation strategy should have the following properties:

$\bullet$ $\mathcal{A}_1$ \emph{(full-coverage)}: the (sub)masks should cover the whole face to avoid blind spots and loss of information.

$\bullet$ $\mathcal{A}_2$ \emph{(no-overlap)}: submasks should not overlap with others, as this introduces to uncertainty with respect to the prototype to be regressed to.

$\bullet$ $\mathcal{A}_3$ \emph{(balanced)}: submasks should have similar sizes.

As can be seen from Table~\ref{tab:ablation_submask_properties}, $\textsc{SM}_{\text{SLADD}}$ has overlapping submasks (marked red) and does not cover the entire face, whereas the mesh-grid strategy has  submasks of unequal size. The grid strategy is the only one satisfying all the above properties, which explains its advantage over the competing submask-generation schemes.
\begin{table}[t]
  \vspace{-3mm}
  \begin{minipage}{.67\linewidth}
    \centering
    \vspace{-2mm}
    \resizebox{1\textwidth}{!}{%
      \begin{tabular}{lrrr|c}\toprule
                                             & $\mathcal{A}_1$ & $\mathcal{A}_2$ & $\mathcal{A}_3$ & Avg.          \\\midrule
        (a) $\textsc{M}_{\text{ConvexHull}}$ & \checkmark      & \checkmark      &                 & 58.5          \\
        \midrule
        (b) $\textsc{SM}_{\text{SLADD}}$     &                 &                 & \checkmark      & 68.3          \\
        (c) $\textsc{SM}_{\text{Meshgrid}}$  & \checkmark      & \checkmark      &                 & 63.8          \\
        (d) $\textsc{SM}_{\text{Grid 4x4}}$  & \checkmark      & \checkmark      & \checkmark      & \textbf{75.9} \\
        \bottomrule
      \end{tabular}}
  \end{minipage}
  \hfill
  \begin{minipage}{.3\linewidth}
    \includegraphics[width=0.45\textwidth,trim = 25mm 15mm 25mm 20mm, clip]{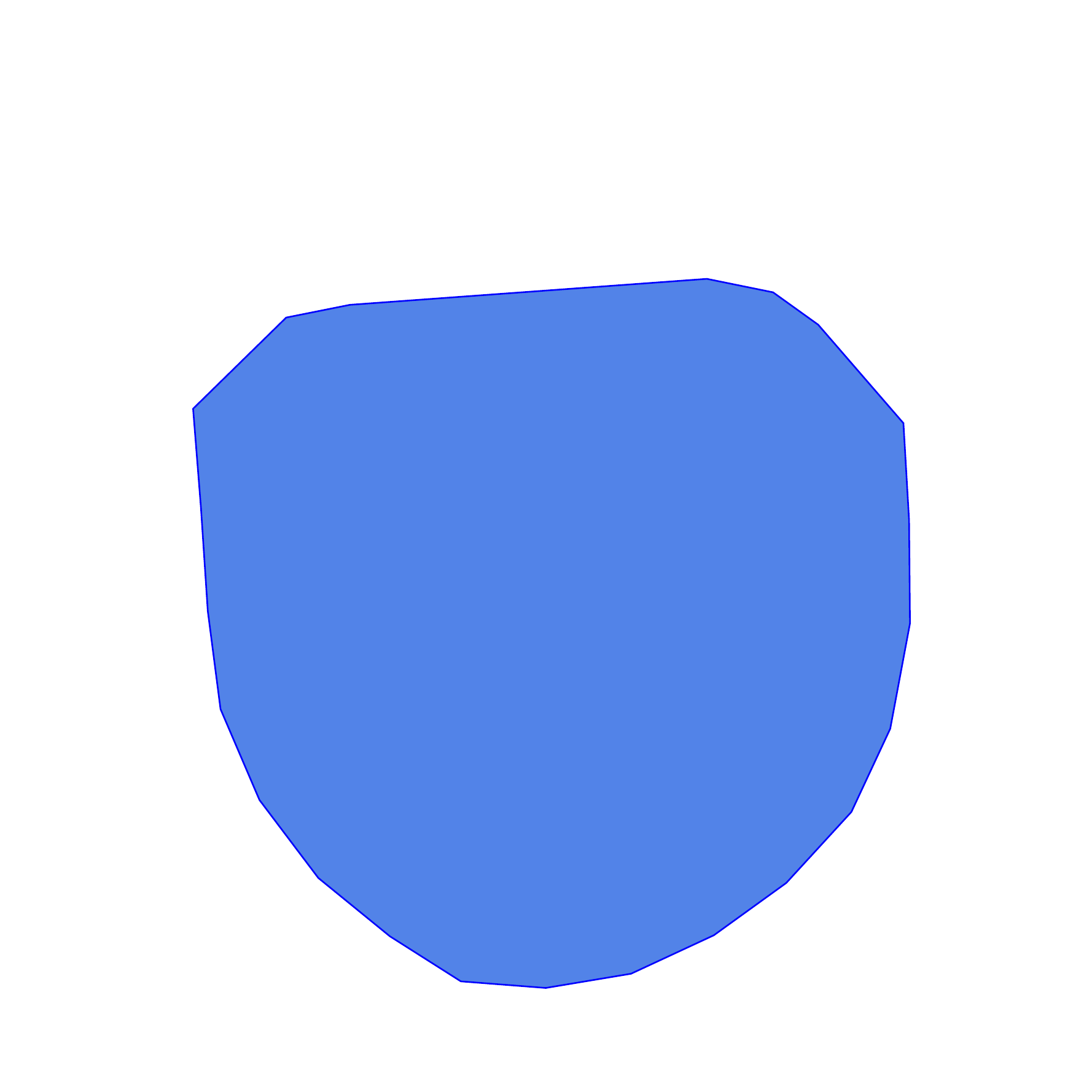}
    \includegraphics[width=0.45\textwidth,trim = 25mm 15mm 25mm 20mm, clip]{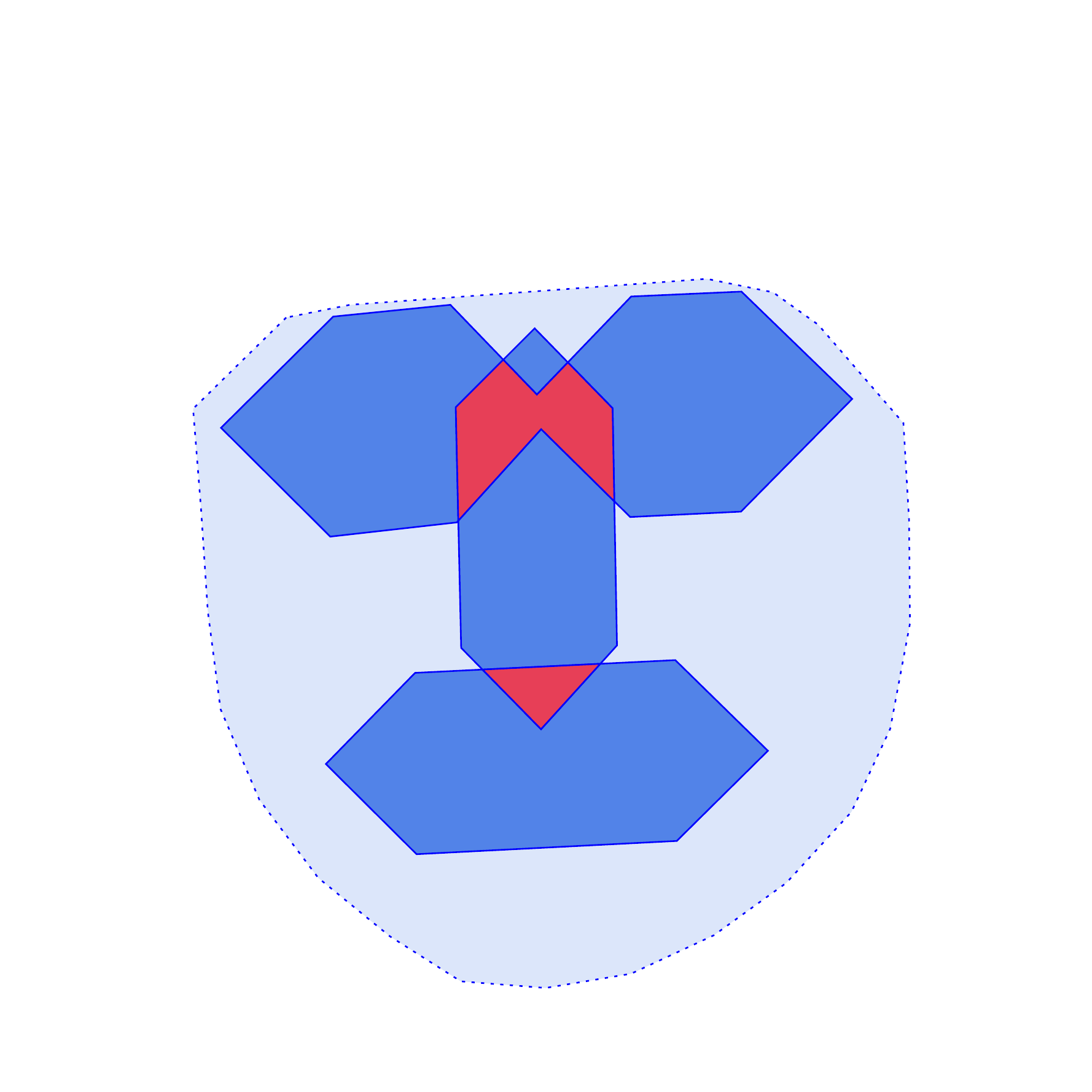}\\ \vspace{-0.1mm}
    {\vspace{-1mm} \footnotesize \hspace{2mm} (a) \hspace{8mm} (b)}\\
    \includegraphics[width=0.45\textwidth,trim = 25mm 15mm 25mm 20mm, clip]{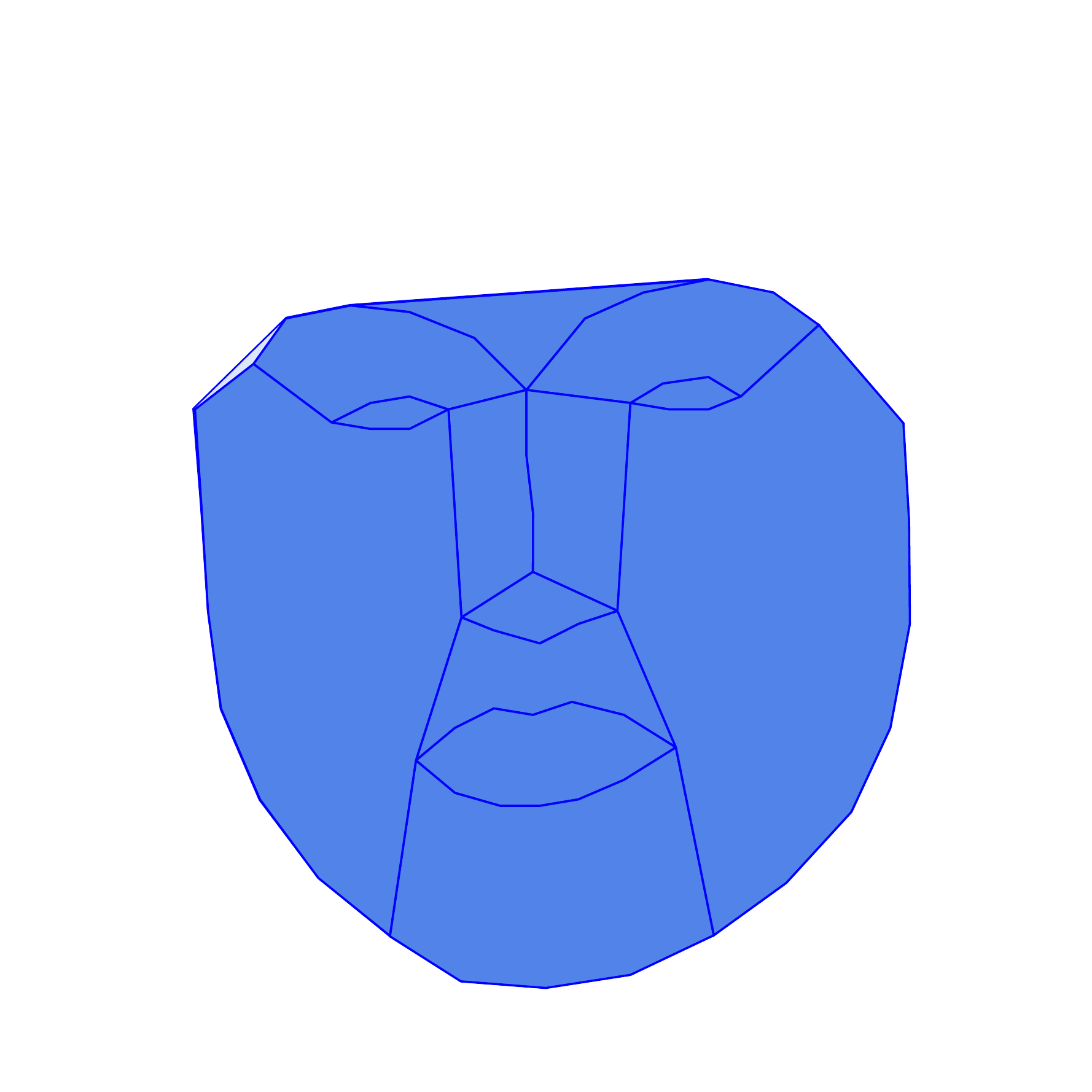}
    \includegraphics[width=0.45\textwidth,trim = 25mm 15mm 25mm 20mm, clip]{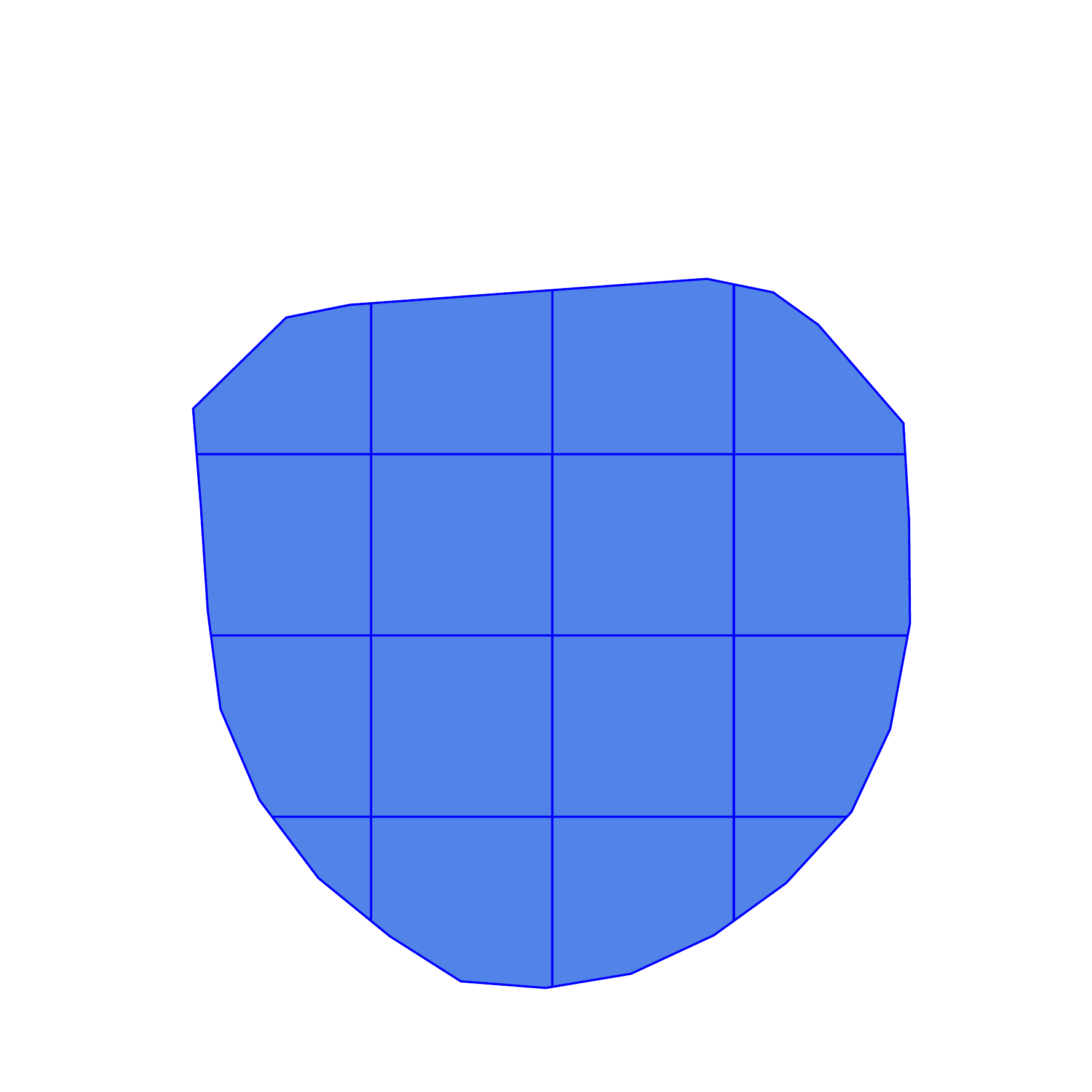}\\ \vspace{-0.1mm}
    {\vspace{-1mm} \footnotesize \hspace{2mm} (c) \hspace{8mm} (d)}\\
  \end{minipage}\vspace{-4mm}
  \caption{\textbf{Performance of different submask schemes.} Results are presented in terms of AUC scores (in \%). The considered submask schemes are shown on the right.\vspace{-1mm}}
  \label{tab:ablation_submask_properties}
\end{table}

\begin{figure}[t]
  \centering
  \includegraphics[width=0.99\columnwidth]{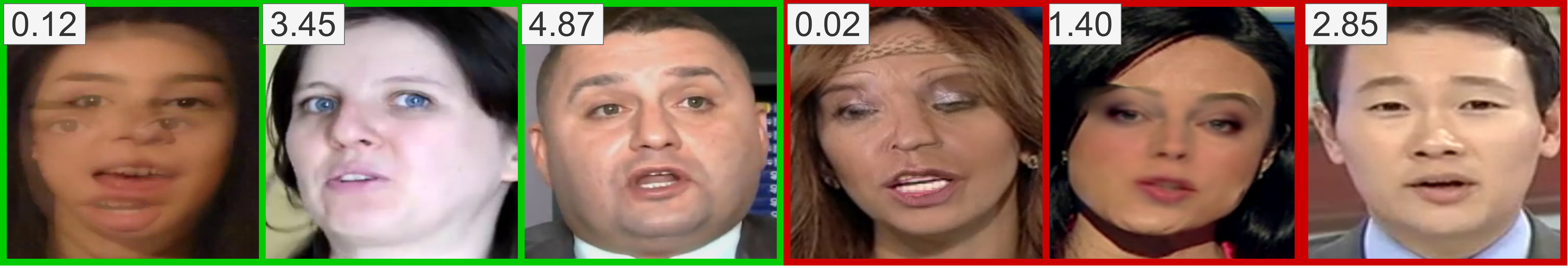}
  \caption{
    \textbf{Visual examples of real (in green) and fake (in red) faces with different anomaly scores.}
  }\vspace{-3mm}
  \label{fig:visual}
\end{figure}

\section{Qualitative analysis}

In Figure \ref{fig:visual}, we show a cross-section of visual examples of real and fake samples with different anomaly scores to analyze the \textbf{strengths and limitations} of SeeABLE. As can be seen, the model performs well overall and generates expected anomaly scores. However, in a small number of cases it also produces low scores for real images that come with deepfake-like artifacts (first image) and  large scores for high-quality deepfakes, such as the one on the far right. %

\section{Conclusion}
In this paper, we presented a powerful new deepfake detector, called SeeABLE, that successfully formalizes deepfake detection as a one-class self-supervised anomaly detection task. The key idea behind the model is to push soft discrepancies synthesized from real faces towards predefined evenly-distributed prototypes using novel learning objectives. The results of our experiments in cross-dataset and cross-manipulation scenarios point to  superior generalizability of SeeABLE over current SoTA methods.
Future work involves improving SeeABLE, \eg, by considering additional losses and pretext tasks. %

\seg{Ethic Statement} We did not identify any potential negative societal impacts of the proposed research. All face images used in this paper were obtained from public datasets. %

\seg{Acknowledgement} Supported by ARRS P2-0250 and P2-0214 and the Eutopia PhD funding scheme.

  {\small
    \bibliographystyle{ieee_fullname}
    \bibliography{egbib}
  }

\appendix
\section*{Appendix}

In the main part of the paper, we introduced SeeABLE, a novel state-of-the-art deepfake detector learned in a one-class learning setting. We evaluated the proposed detector in rigorous experiments in cross-dataset and cross-manipulation scenarios over multiple datasets and in comparison to $12$ state-of-the-art competitors, and presented a number of ablation studies to demonstrate the impact of various model components. In this supplementary material, we now provide: $(i)$ additional technical details on SeeABLE, related to: (a) the definition of the evenly distributed prototype used with the Bounded Contrastive Regression (BCR), and (b) the geometric constraints used with the auxilary guidance loss, %
$(ii)$ visual examples of the generated local image perturbations (i.e., the soft discrepancies) in the spatial and frequency domain, $(iii)$ additional ablations, $(iv)$ qualitative results with examples of face images generated diffusion-based (generative) models, and $(v)$ information on the reproducibility of SeeABLE with  links to relevant (open-access) repositories. %

\section{Hard-prototype generation}

The main idea behind SeeABLE is to generate local image perturbations (soft discrepancies) and then map the different perturbations to a set of so-called \textit{hard-prototypes} that can later be used to derive an anomaly score for deepfake detection. One of the key components in this framework are the hard-protoypes, which are defined in a way that ensures that they are evenly (i.e., equidistantly) distributed on an $n$-dimensional hypercube. This setup not only results in an optimal separability between the different prototypes (in terms of average between prototype distance), but also leads to highly desirable characteristics when used with contrastive learning objectives, as theoretically and empirically demonstrated in \cite{dissectingsupcon}.

In SeeABLE, we compute the hard prototypes in accordance with the algorithm from \cite{MMAlgo}. Here, the $n$-dimensional prototypes $\{\boldsymbol{p}_1 \ldots \boldsymbol{p}_K\}$, for $K \in [2, \ldots, n+1]$, which serve as the targeted optimal representation of the generated soft discrepancies, are defined as vertices of a regular simplex, i.e.:
\begin{equation}
  \boldsymbol{p}_{i} =
  \begin{cases}
    \frac{1}{\sqrt{n}} \mathbf{1},                                                            & \text{if } K = 1                  \\
    -\frac{1+\sqrt{n+1}}{n^{\frac{3}{2}}} \mathbf{1} + \sqrt{\frac{n+1}{n}} \mathbf{e}_{i-1}, & \text{if } K \in [2, \ldots, n+1]
  \end{cases},
\end{equation}
where $\boldsymbol{1} \in \mathbb{R}^{n}$ is a vector of all ones and $\mathbf{e}_{i-1}$ is a one-hot encoded vector of all zeros with a `$1$' at the $(i-1)^{th}$ position, and $i=1,\ldots, K$. %

\section{Guidance loss and geometric contraints}

\begin{figure}[h]
  \centering
  \includegraphics[width=0.33\linewidth, trim = 8mm 2mm 4mm 10mm, clip]{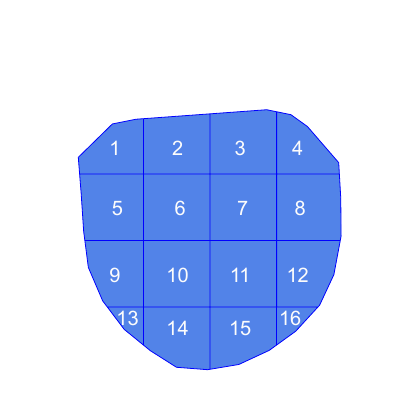}
  \hspace{4mm}
  \includegraphics[width=0.33\linewidth, trim = 8mm 2mm 4mm 10mm, clip]{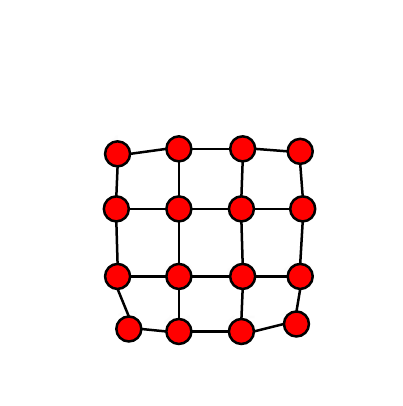}\vspace{2mm}
  \caption{\textbf{Illustration of the $4\times 4$ grid-based submask generation scheme and corresponding graph.} The graph definition on the right is used to define the graph-based distance used for the definition of the geometric constraint and respective guidance loss.}
  \label{fig:my_label_example}
\end{figure}

SeeABLE is learned by minimizing a learning objective that consists of a weighted combination of the proposed Bounded Contrastive Regression (BCR) loss and a \textit{Guidance loss} that encourages the model to localize the generated soft-discrepancies, while taking \textit{geometric constraints} into account. Specifically, the guidance loss, defined in Eqs. (12) and (13) of the main paper, uses a three-scale penalty definition for the learning procedure: (1) the lowest penalty of $2^{-2}$ is assigned if the predicted and true location of the soft-discrepancy overlap, (2) a higher penalty of $2^{-1}$ is assigned if the predicted and true location stem from a (horizontally) mirror-symmetric region of the face (e.g., regions $6$ and $7$ in Fig.~\ref{fig:my_label_example}, or regions $9$ and $12$, etc.), and (3) the highest penalty, proportional to the graph-distance $d_{graph}$ between the predicted and true soft-discrepancy location is assigned in all other cases, i.e., $2^0\times d_{graph}$.

To define the graph-based distance $d_{graph}$ for the guidance loss, we transform the submask scheme into a graph representation, as illustrated in Fig. \ref{fig:my_label_example} for a $4 \times 4$ grid strategy. Here, each of the $N_{loc}$ patches is represented by a red node, with two nodes being connected by an edge if they share a common border. An adjacency matrix $\notvector{A}$ can then be constructed from this graph, where $\notvector{A}[i, j] = d$ if node $i$ is connected to node $j$, and $\notvector{A}[i, j] = 0$ otherwise, with $d$ representing the length of the edge between $i$ and $j$. However, in SeeABLE, we are not concerned with the edge length and fix it to $d=1$. The graph-based distance $d_{graph}$ is thus defined as the total (minimum) number of edges that need to be traversed to connect nodes $i$ and $j$. We note that this definition is also applicable to other competing submask-generation schemes considered in the main part of the paper.

\section{Examples of soft-discrepancies}

\begin{figure}[t!]
  \includegraphics[width=1.\linewidth]{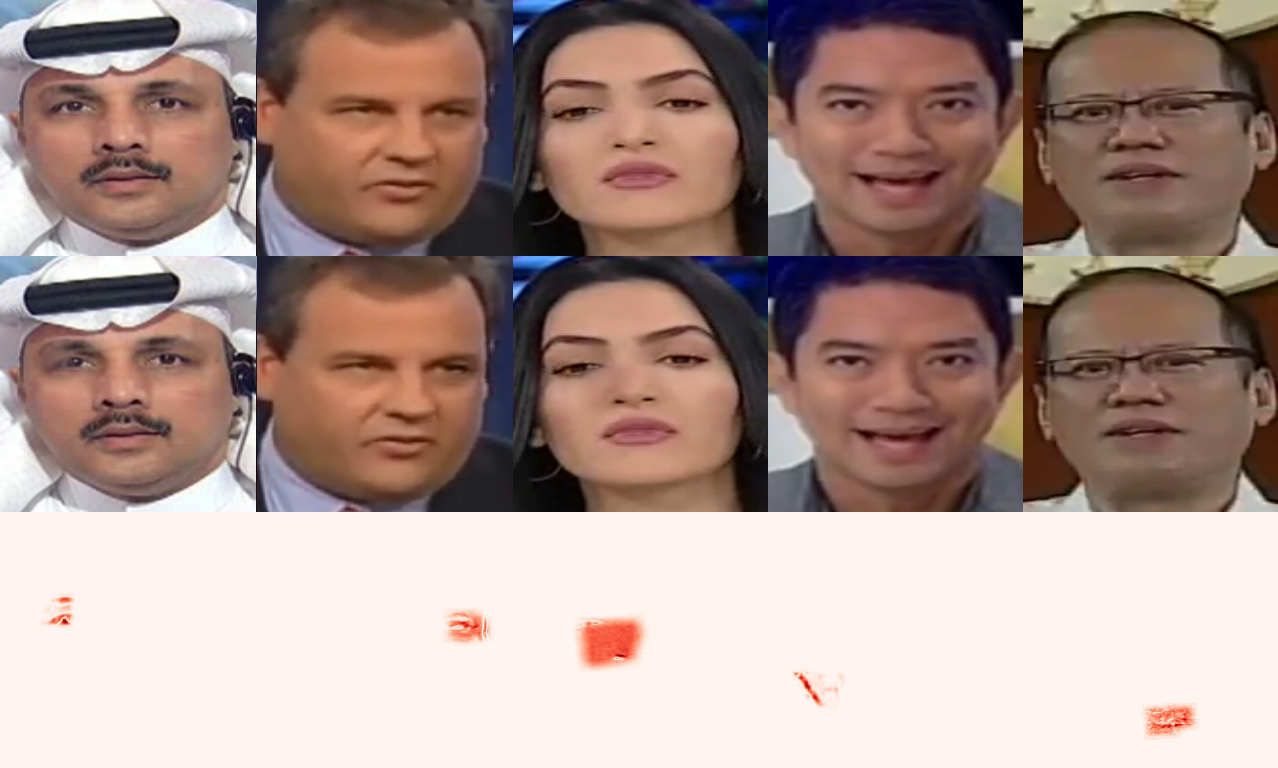}\vspace{2mm}
  \caption{\textbf{Illustration of the visual impact of the generated soft discrepancies.} The first row shows examples of real faces, the second row shows the perturbed versions and the last row shows their absolute differences.} %
  \label{fig:discrepancy_diff}
\end{figure}

\begin{figure*}[!t]
  \centering
  \includegraphics[width=\linewidth]{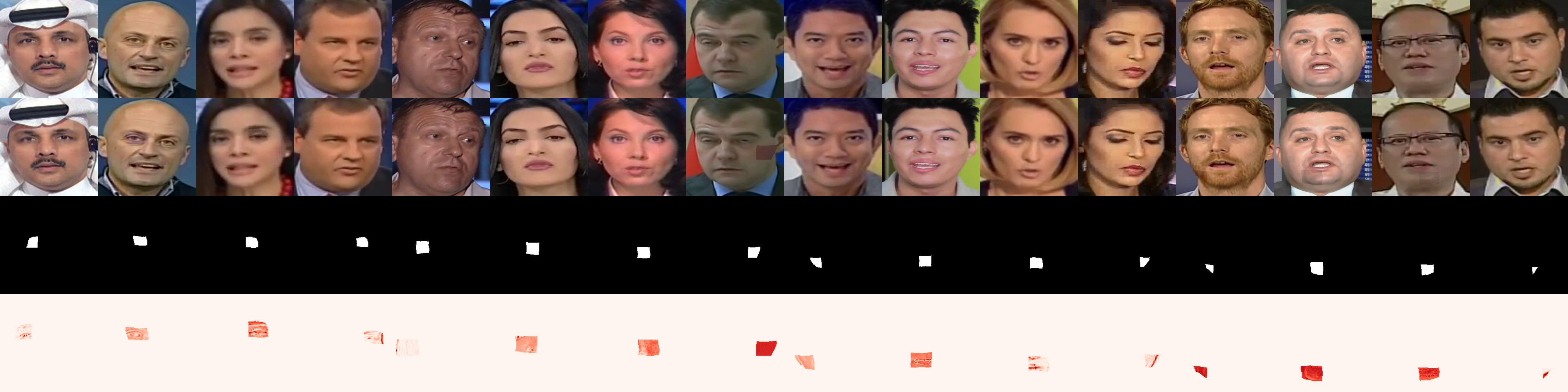}\vspace{2mm}
  \caption{\textbf{Impact of the soft discrepancies on the visual appearance of the perturbed faces across a diverse set of examples.} The first and second row show the real and locally perturbed faces, respectively. The third row shows examples of the blending masks generated by the grid-based submask-generation scheme, and the last row shows the absolute differences between the initial and perturbed faces. %
  }
  \label{fig:submasks_partition}
\end{figure*}

To illustrate the impact of the generated soft discrepancies on the visual appearance of the perturbed faces, we show in
Figure \ref{fig:discrepancy_diff} a number of examples. Here, the first row presents the original images, the second row shows the locally perturbed faces and the last row shows the absolute difference between the two. Note how the soft discrepancies are hardly visible, but still allows learning a highly capable deepfake detector. In Figure \ref{fig:submasks_partition}, we present additional examples across a wider and more diverse set of images, but in addition to the real and perturbed faces and their difference, we also show the blending masks in the third row. Observe how different local areas of the face are targeted by the soft discrepancies, leading to subtle perturbations that are often imperceptible to the human visual system but detectable by SeeABLE.

\section{Additional ablations}

In the main part of the paper, we show several ablation studies to explore the impact of the different components of SeeABLE on the detection performance. In this section, we now add to these results with two additional ablation experiments that investigate: $(i)$ the impact of spatial and frequency-domain perturbation on the detection task, and $(ii)$ the effect of different grid configurations in the submask-generation scheme.

\subsection{Spectral vs. spatial perturbations}

Let the complete set of local perturbations, utilized to generate the soft-discrepancies for SeeABLE, be denoted as $\mathcal{P}$ and let this set consist of perturbations being applied in either the spatial or the frequency domain, i.e., $\mathcal{P}=\{\mathcal{P}_{spatial}, \mathcal{P}_{freq}\}$. The set of perturbations (transformations) used in either domain was defined in the main part of the paper in Section 4.1. In
Table \ref{tab:freq_spatial}, we investigate the impact of each group of perturbations on the detection performance of SeeABLE on FF+ HQ. The AUC score is again reported as a performance indicator. %
\setcounter{table}{7}
\begin{table}[h]
  \centering
  \resizebox{0.85\columnwidth}{!}{%
    \begin{tabular}{l|cccc|c}
      \toprule
      \multirow{2}{*}{$\mathcal{P}$}                  & \multicolumn{5}{c}{Test set - AUC (in \%)}                                                                 \\
      \cmidrule{2 - 6}                                & DF                                         & F2F           & FS            & NT            & Avg.          \\
      \midrule
      $\{\mathcal{P}_{spatial}\}$                     & 96.4                                       & 94.1          & 95.8          & 94.8          & 95.3          \\
      $\{\mathcal{P}_{freq}\}$                        & 99.6                                       & 97.3          & 96.9          & 93.2          & 96.7          \\
      \midrule
      $\{\mathcal{P}_{spatial}, \mathcal{P}_{freq}\}$ & \textbf{99.2}                              & \textbf{98.8} & \textbf{99.1} & \textbf{96.9} & \textbf{98.5} \\
      \bottomrule
    \end{tabular}}\vspace{3mm}
  \caption{\textbf{Ablation results with respect to the augmentation type used.} Shown are AUC scores (in \%) on the FF++ HQ dataset.}
  \label{tab:freq_spatial}
\end{table}

As can be seen, the results clearly show that the different perturbation types are complementary to each other. By combining spatial and frequency-domain perturbations, SeeABLE obtains better results than with either of alone. We note again at this point that the number and type of perturbations considered during training ($N_{type}$) defines the number of prototypes used for the regression task of SeeABLE. In turn, the results in Table \ref{tab:freq_spatial} also illustrate the impact of changing the number of prototypes when learning the detection model.  %

\subsection{Sensitivity to the grid size}

In the main part of the paper, we showed that the grid-based strategy to submask generation yielded the best performance among the evaluated schemes. In Table \ref{tab:ablation_submask_properties}. we now explore the impact of different configurations of this grid. Specifically, we investigate the use of  3$\times$3, 4$\times$4, and 5$\times$5 grids in SeeABLE and observe strong performance across all of these configurations with the highest results observed with the $4 \times 4$ configuration. This particular configuration appears to offer a good trade-off between locality and visual-perturbation-impact to facilitate learning a well performing detection model. We note that the  $\mathcal{A}$ notation stands for the mask criteria defined in the main part of the paper, i.e., $\mathcal{A}_1$ (full coverage), $\mathcal{A}_2$ (no overlap), $\mathcal{A}_3$ (balanced size).

\begin{table}[h]
  \centering
  \resizebox{0.8\columnwidth}{!}{%
    \begin{tabular}{lrrr|c}\toprule
                                      & $\mathcal{A}_1$ & $\mathcal{A}_2$ & $\mathcal{A}_3$ & Avg. - AUC (in \%) \\\midrule
      $\textsc{SM}_{\text{Grid 3x3}}$ & \checkmark      & \checkmark      & \checkmark      & 74.9               \\
      $\textsc{SM}_{\text{Grid 4x4}}$ & \checkmark      & \checkmark      & \checkmark      & \textbf{75.9}      \\
      $\textsc{SM}_{\text{Grid 5x5}}$ & \checkmark      & \checkmark      & \checkmark      & 75.1               \\
      \bottomrule
    \end{tabular}}\vspace{3mm}
  \caption{\textbf{Impact of different grid configurations on the performance of SeeABLE on the DFDC dataset.} Results are shown for different submask-generation strategies, with the $4 \times 4$ grid performing the best among the tested configurations.}
  \label{tab:ablation_submask_properties}
\end{table}

\section{Diffusion models}

\begin{figure}
  \centering
  \includegraphics[width=0.99\columnwidth] {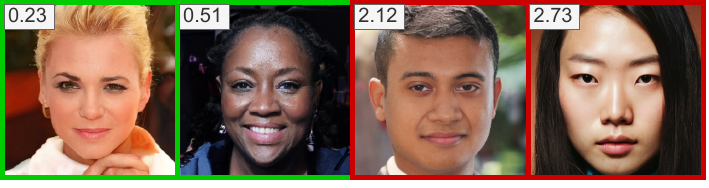}\vspace{1mm}
  \caption{\textbf{Visual examples of randomly selected real (in green) and diffusion-generated  faces (in red) with corresponding anomaly scores.} The first two images represent real faces from CelebA-HQ \cite{karras2017progressive} and FFHQ \cite{karras2019style}, respectively, the third image was generated with latent diffusion \cite{rombach2022high} and the fourth with Midjourney \cite{midjurney}.}
  \label{fig:visual}
\end{figure}

The recent proliferation of probabilistic diffusion models has led to the creation of numerous synthetic image datasets \cite{croitoru2022diffusion,yang2022diffusion}. However, there remains a dearth of facial datasets specifically designed for deepfake detection. Additionally,  techniques employed for producing high-quality, non-existent facial images with a high degree of realism fall under the umbrella of \textit{entire face synthesis} rather than deepfake generation. Such techniques predominantly utilized Generative Adversarial Networks (GANs), but are now increasingly adopting denoising diffusion probabilistic models for the synthesis task.

Similarly to the majority of work on deepfake detection available in the literature, our paper focuses on the detection of input-conditioned face manipulations, where local regions of the original faces are altered, and not synthesis procedures that generate entire (artificial) face images. %
Although the intricate details of (entire) face synthesis techniques are beyond the scope of this paper, we showcase the effectiveness of SeeABLE for the detection of synthetically generated full face images in Figure \ref{fig:visual}. Here, we apply our model to two real face images from the CelebA-HQ \cite{karras2017progressive} and FFHQ \cite{karras2019style} datasets and two diffusion- diffusion-generated images - the first generated with latent diffusion \cite{rombach2022high} and the second with Midjourney \cite{midjurney}. As can be seen from the reported anomaly scores in the corners of the presented images, SeeABLE ensures good separation between the real and synthetic images and produces comparably higher anomaly scores for the synthesized faces, despite the fact that it was not trained specifically for the detection of such types of data. %

\section{Reproducibility}

We note that all of our experiments are fully reproducible. The source code, training scripts, models and learned (model) weights, associated with SeeABLE, are made publicly available here:
\begin{itemize}
  \item SeeABLE: \\
        {\footnotesize \url{https://github.com/anonymous-author-sub/seeable}}
\end{itemize}

The remaining code used in the paper is also available from the official repositories, i.e.,
\begin{itemize}
  \item Dlib: \\
        {\footnotesize \url{http://dlib.net/}}
  \item RetinaFace: \\
        {\footnotesize \url{https://github.com/deepinsight/insightface}}
  \item FaceSwap:\\
        {\footnotesize \url{www.github.com/deepfakes/faceswap}}

\end{itemize}

\end{document}

%% file: notation.tex
\usepackage{mathtools} %

\DeclarePairedDelimiter\abs{\lvert}{\rvert}
\DeclarePairedDelimiter\norm{\lVert}{\rVert}

\DeclarePairedDelimiter\ensemble{\lbrace}{\rbrace}

\DeclareMathOperator*{\argmin}{arg\,min}

\newcommand{\seg}[1]{\vspace{0.3mm} \noindent\textbf{#1}.}
\newcommand{\isep}{\mathrel{{.}\,{.}}\nobreak}
\newcommand{\secref}[1]{\S\ref{#1}}

\newcommand{\notloss}[1]{\mathcal{L}_{\text{#1}}}

\newcommand{\notvector}[1]{\mathbf{#1}}

\newcommand{\notfunction}[1]{\mathit{#1}}

\newcommand{\notgt}[1]{\tilde{#1}}

\newcommand{\notintinter}[1]{[1\mathrel{{.}\,{.}}\nobreak#1]}